\documentclass[conference]{IEEEtran}
\IEEEoverridecommandlockouts
\usepackage{cite}
\usepackage{amsmath,amssymb,amsfonts}
\usepackage{algorithmic}
\usepackage{graphicx}
\usepackage{textcomp}
\usepackage{xcolor}
\usepackage{subcaption}
\def\BibTeX{{\rm B\kern-.05em{\sc i\kern-.025em b}\kern-.08em
    T\kern-.1667em\lower.7ex\hbox{E}\kern-.125emX}}
\begin{document}

\title{Wireless Center of Pressure Feedback System for Humanoid Robot Balance Control using ESP32-C3 \\
}

\author{%
Muhtadin$^{1}$$^{,}$$^{3}$,
Faris Rafi Pramana$^{2}$,
Dion Hayu Fandiantoro$^{1}$$^{,}$$^{2}$, Moh Ismarintan Zazuli $^{4}$, Atar Fuady Babgei $^{1}$$^{,}$$^{5}$ \\

\small $^{1}$Department of Computer Engineering, Institut Teknologi Sepuluh Nopember, Surabaya, Indonesia \\
\small $^{2}$Graduate School of Science and Technology, Kumamoto University, Japan \\
\small $^{3}$The Science and Technology Center of Artificial Intelligence for Healthcare and Society (PUI AIHeS), Indonesia \\
\small $^{4}$Department of Electrical Automation Engineering, Institut Teknologi Sepuluh Nopember, Surabaya, Indonesia \\
\small $^{5}$Department of Computing, Imperial College London, United Kingdom \\
Email: muhtadin@its.ac.id,  farisrafp@gmail.com, dion.fandiantoro@st.cs.kumamoto-u.ac.jp, ismarintan@its.ac.id, atarbabgei@its.ac.id.
}

\maketitle

\begin{abstract}
Maintaining stability during the single-support phase is a fundamental challenge in humanoid robotics, particularly in dance robots that require complex maneuvers and high mechanical freedom. Traditional tethered sensor configurations often restrict joint movement and introduce mechanical noises. This study proposes a wireless embedded balance system designed to maintain stability on uneven surfaces. The system utilizes a custom-designed foot unit integrated with four load cells and an ESP32-C3 microcontroller to estimate the Center of Pressure (CoP) in real time. The CoP data were transmitted wirelessly to the main controller to minimize the wiring complexity of the 29-DoF VI-ROSE humanoid robot. A PID control strategy is implemented to adjust the torso, hip, and ankle roll joints based on CoP feedback. Experimental characterization demonstrated high sensor precision with an average measurement error of 14.8 g. Furthermore, the proposed control system achieved a 100\% success rate in maintaining balance during single-leg lifting tasks at a 3-degree inclination with optimized PID parameters (Kp=0.10, Kd=0.005). These results validate the efficacy of wireless CoP feedback in enhancing the postural stability of humanoid robots, without compromising their mechanical flexibility.
\end{abstract}


\section{Introduction}
In recent years, rapid advancements in robotics have increased human–robot interactions, particularly in the context of humanoid robots. Maintaining balance and stability is a key aspect of the design of humanoid robots\cite{Chiang}. This is particularly important in traditional dance robot competitions, where robots are not allowed to fall while crossing the arena and must be able to perform dances with complex movements.

Recent studies on humanoid robot balance have focused on creating control systems with advanced sensors, such as gyroscopes and accelerometers \cite{Mohammadi,Zhang,Zhao}. These systems use intelligent algorithms to mimic human stability. However, they struggle to accurately detect ground forces, especially on uneven surfaces or when disturbed. This is because inertial sensors cannot directly measure how weight is distributed, and the algorithms require a large amount of training data in controlled settings. This renders robots less effective in real-world situations. Using load cells on the robot's feet is a good solution for this problem. They can measure pressure and weight distribution in real time, providing important data to help the robot adjust its position and movement quickly \cite{Farraj,Liu,Kim}. This works well with inertial sensors to create a better and more reliable balance system for the user.

Ilyasaa \cite{Ilyasaa} discussed load cell sensors on humanoid robots with 24 degrees of freedom for balance control while walking. Muhtadin \cite{muhtadin} implemented a load cell in a humanoid soccer-robot to detect foot pressure and regulate step pitch parameters through PID control. The robot maintained balance at 0°-25° tilt angles but fell at 30°. While PID control proved effective, the research lacked an analysis of the impact of tilt angles during walking. 

Wulandari \cite{Wulandari} implemented Load Cell sensors, Kalman Filter, and PID Controller for balancing a dancing humanoid robot for the Indonesian Robot Dance Competition (KRSTI). Using MPU6050 and Load Cell sensors for Center of Pressure detection, the robot achieved success rates of 87.5\% and 89\% in standing and dancing, respectively. Although sensors and algorithms were effectively combined, testing was limited to the double-support phase, and cable-based hardware restricted foot movements.

We have a humanoid robot (VI-ROSE ITS) that performs traditional Indonesian dances in the KRSTI competition. The robot occasionally fell while dancing, particularly when walking or lifting one foot, and costume accessories further challenged its balance. Load cell sensors can provide information about the robot's center of pressure, detect detailed pressure changes to measure balance changes precisely, and respond rapidly. Wireless radio signals have been proposed to improve the data transmission efficiency. This study aims to develop an embedded system on the soles of a 29-DoF humanoid dance robot to read the pressure center using load cell sensors and control servos for balance on inclined surfaces. The testing used static motion to lift either foot, with servos controlling the roll position. This paper covers the design and implementation of load cell sensors, PID control system, sensor reading results, and PID parameters' effect on balance, and concludes with findings and future research suggestions.

\section{Design and Implementation}
\label{sec:designandimplementation}

\subsection{System Block Diagram}
    \label{subsec:systemblockdiagram}

    \begin{figure} [bt] \centering
        \includegraphics[width=0.48\textwidth]{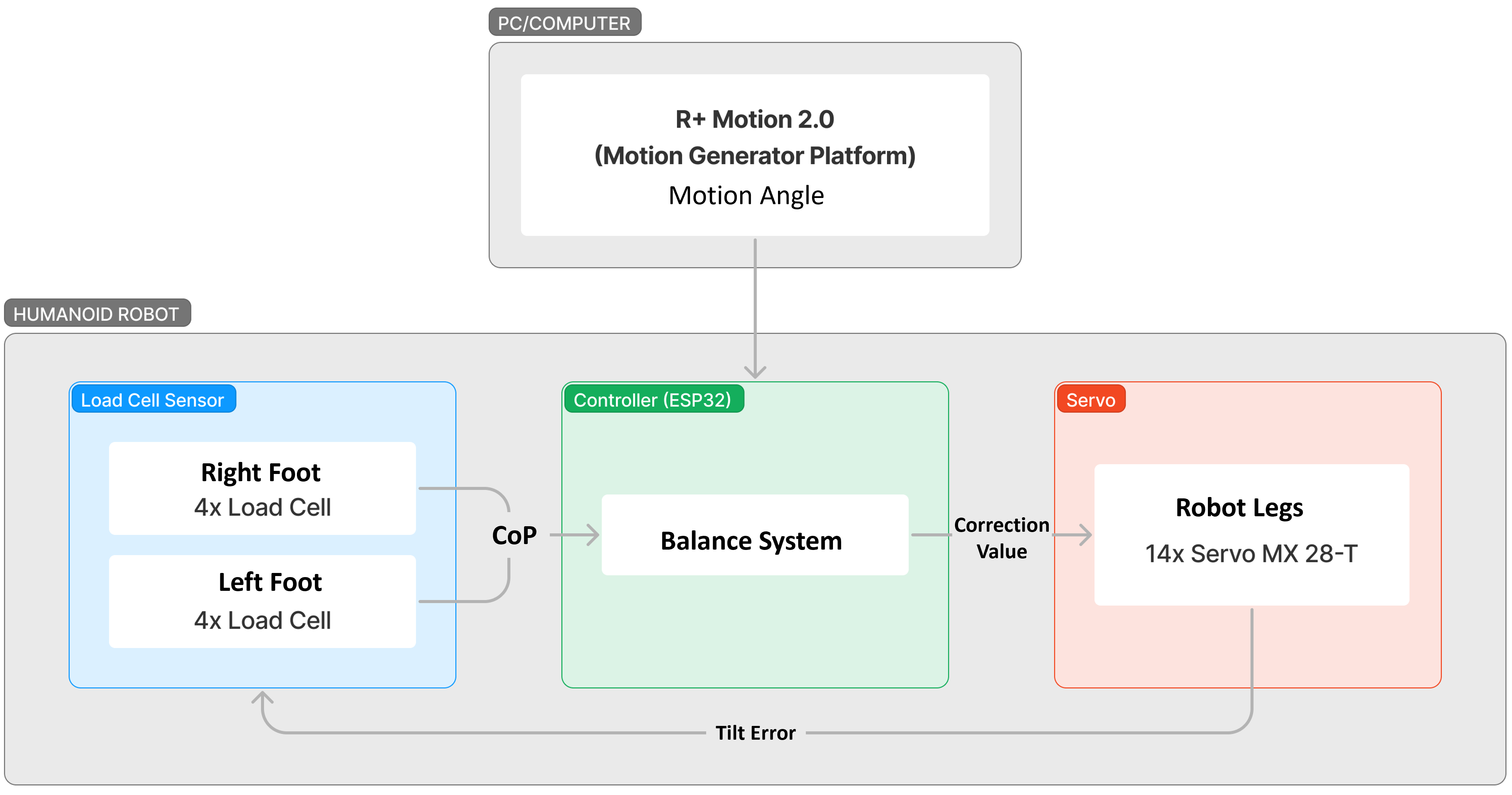}
        \caption{Overall System Diagram}
        \label{fig:Diagram_Sistem}
    \end{figure}
    
    The sensor part includes load cells mounted on each foot of the robot. These load cells detect the load or pressure on the feet of the robot. The control part involves a microcontroller responsible for processing the data from the sensors and controlling the robot's movements. The mechanical system includes the robot frame, which consists of various servos used to move the robot. The motion data part stores pre-designed movements in the microcontroller file system. These data consist of several frames containing an array of target positions for each servo along with the time required to reach those positions. The system diagram is shown in Figure \ref{fig:Diagram_Sistem}.
    
\subsection{Mechanical System}
    \label{subsec:mechanicalsystem}

\begin{figure} [bt] \centering
      \includegraphics[width=0.5\textwidth]{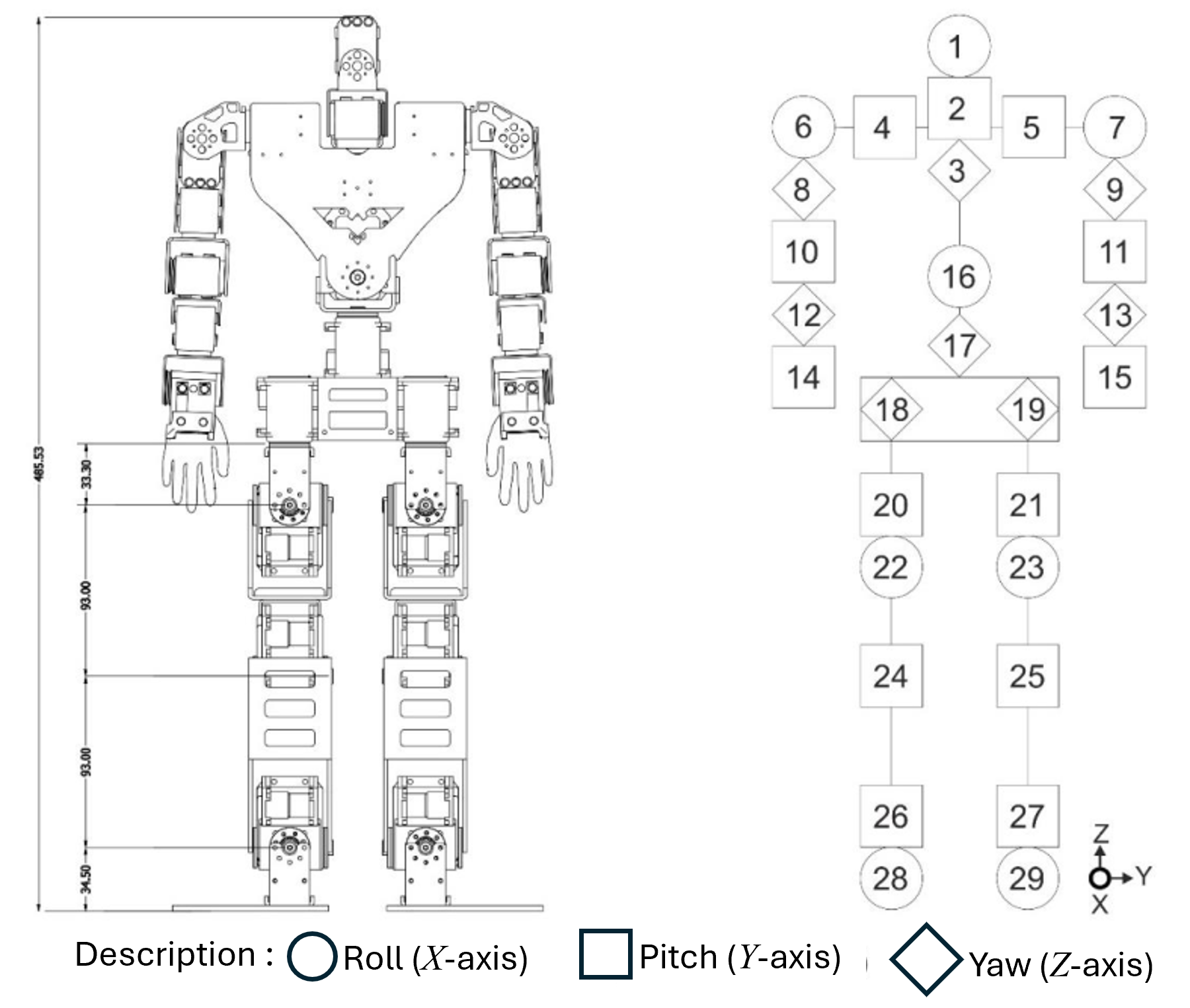}
      \caption{Mechanical Design of the Robot and Servo ID Naming}
      \label{fig:Desain_Mekanik}
    \end{figure}

   The VI-ROSE robot was developed based on the ICHIRO \cite{ichiro_winning} humanoid soccer robot. Additional motors were integrated into the hips and arms to facilitate smooth movements while performing traditional dances. The humanoid robot design encompasses 29 degrees of freedom (DoFs). The upper and lower bodies were equipped with 15 XL-320 servos and 14 MX-28 servos, respectively. We employed a communication system in conjunction with Dynamixel motors \cite{dynamixel_system}, manufactured by Robotis, South Korea, to ensure precise and efficient control in our application. The detailed design, dimensions, and unique servo ID of the robot are shown in Figure \ref{fig:Desain_Mekanik}.. 

\subsection{Electronic System}
    \label{subsec:electronicsystem}

\begin{figure} [bt] \centering
      \includegraphics[width=0.4\textwidth]{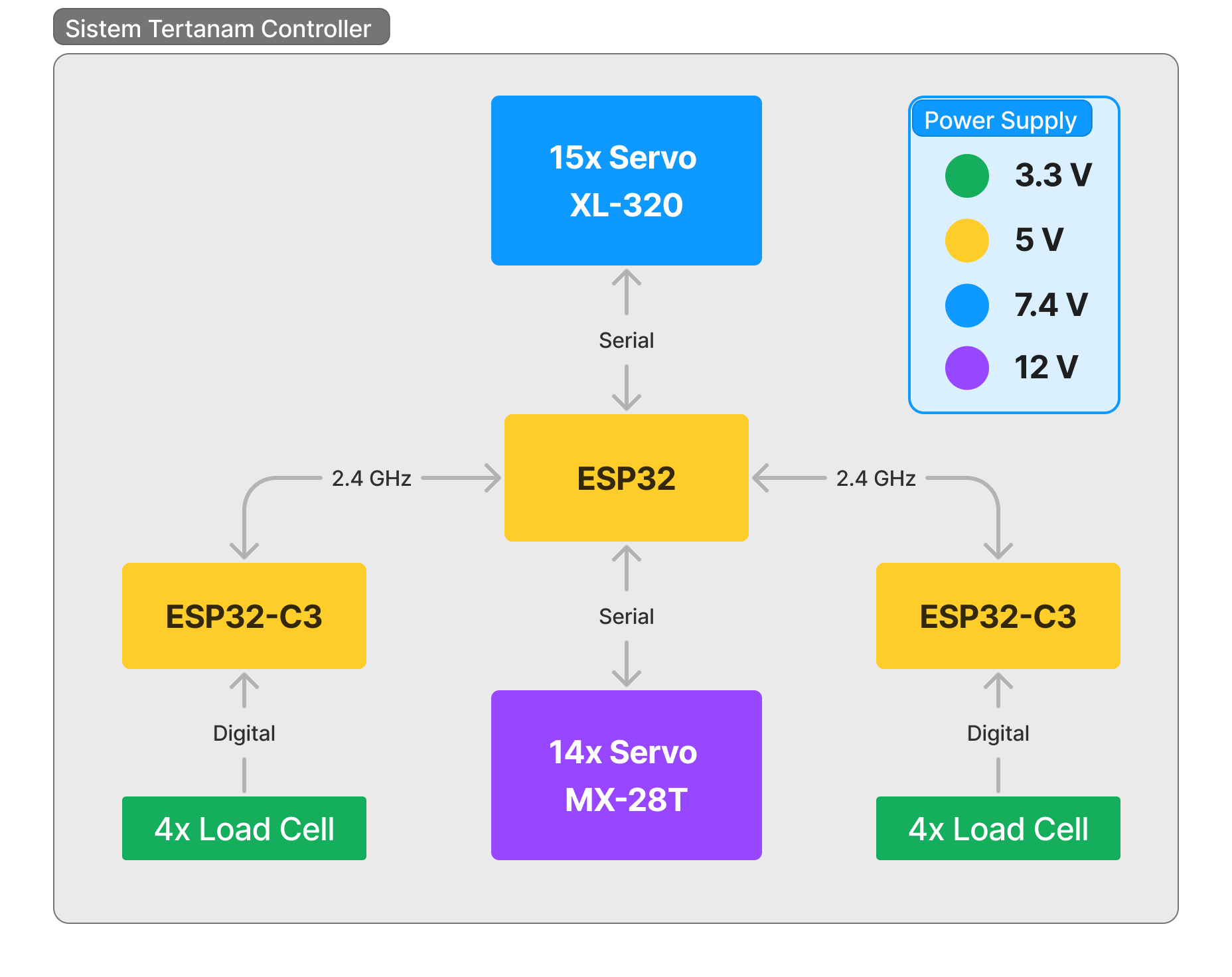}
      \caption{Electronic and Communication Diagram Between Components}
      \label{fig:Diagram_Elektronik}
    \end{figure}

    The hardware system used in this study is explained using the block diagram shown in Figure \ref{fig:Diagram_Elektronik}. This system uses an embedded system consisting of ESP32 and ESP32-C3 microcontroller units. The embedded system was chosen because the robot developed in this study is an improvement over the previous research conducted by Fahd (2018)\cite{fahd}.

    ESP32-C3 is used for data acquisition from the load cell and sends it to the ESP32. The ESP32-C3 has the same Wi-Fi capability as the ESP32, allowing wireless communication between the two microcontrollers. The ESP32-C3 also has a smaller dimension, making it easier to place on the robot's foot.

\subsection{Foot Design of the Robot}
    \label{subsec:desaignsystemloadcell}

\begin{figure} [bt] \centering
      \includegraphics[width=0.4\textwidth]{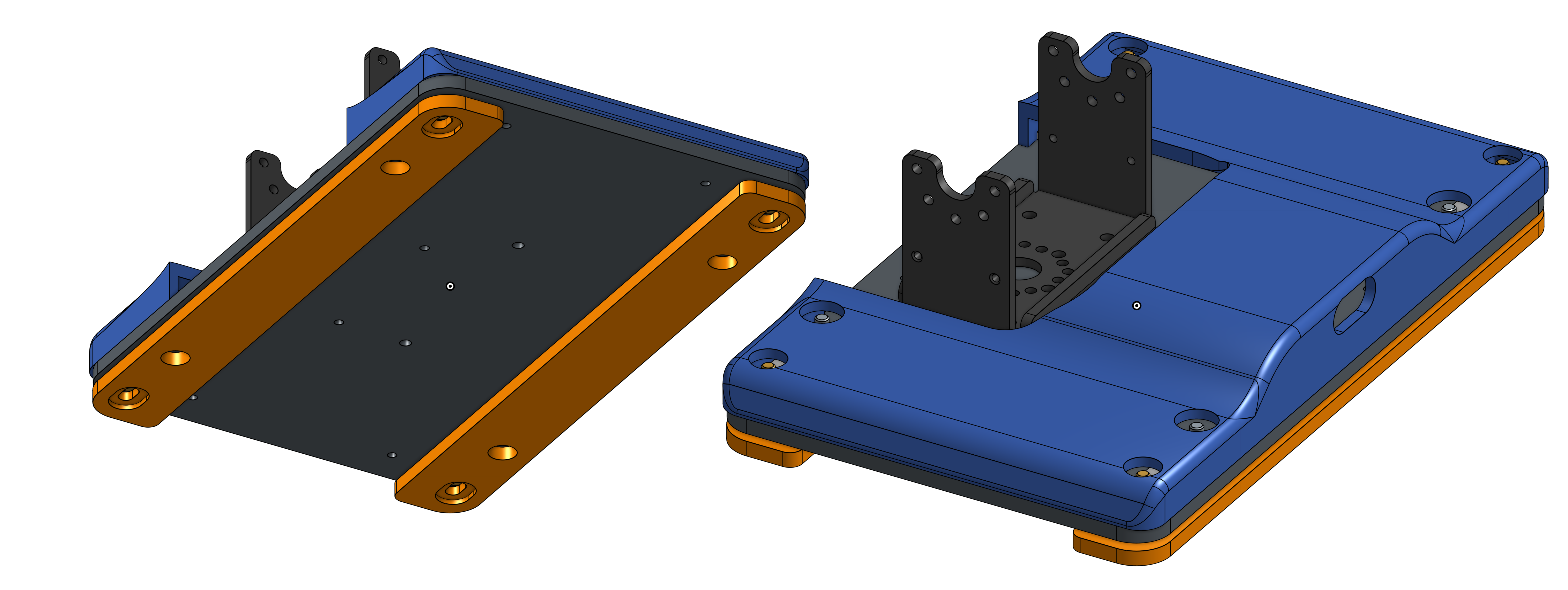}
      \caption{Overall Design of the Robot Foot}
      \label{fig:Desain_Kaki}
    \end{figure}

    \hspace*{1em} Each robot foot is equipped with 4 load cells mounted at the ends of the foot. Each load cell detects the pressure, allowing the system to determine the center of pressure on the robot's foot. The design of the robot foot is shown in Figure \ref{fig:Desain_Kaki}.
    
    \hspace*{1em} In Figure \ref{fig:Desain_Kaki}, it is shown that each load cell is placed at the end of the robot's foot. The load cells were connected to a microcontroller located inside the robot's foot. For the electronics, a closure made from 3D printing was provided to protect the components from damage. For the mechanical part, the load cell sensors were equipped with pads that served as pressure points for the load cells and provided anti-slip to prevent slipping.

\subsection{Load Cell Coefficient Configuration}
    \label{subsec:configurationcoefficient}

\begin{figure} [bt] \centering
      \includegraphics[width=0.4\textwidth]{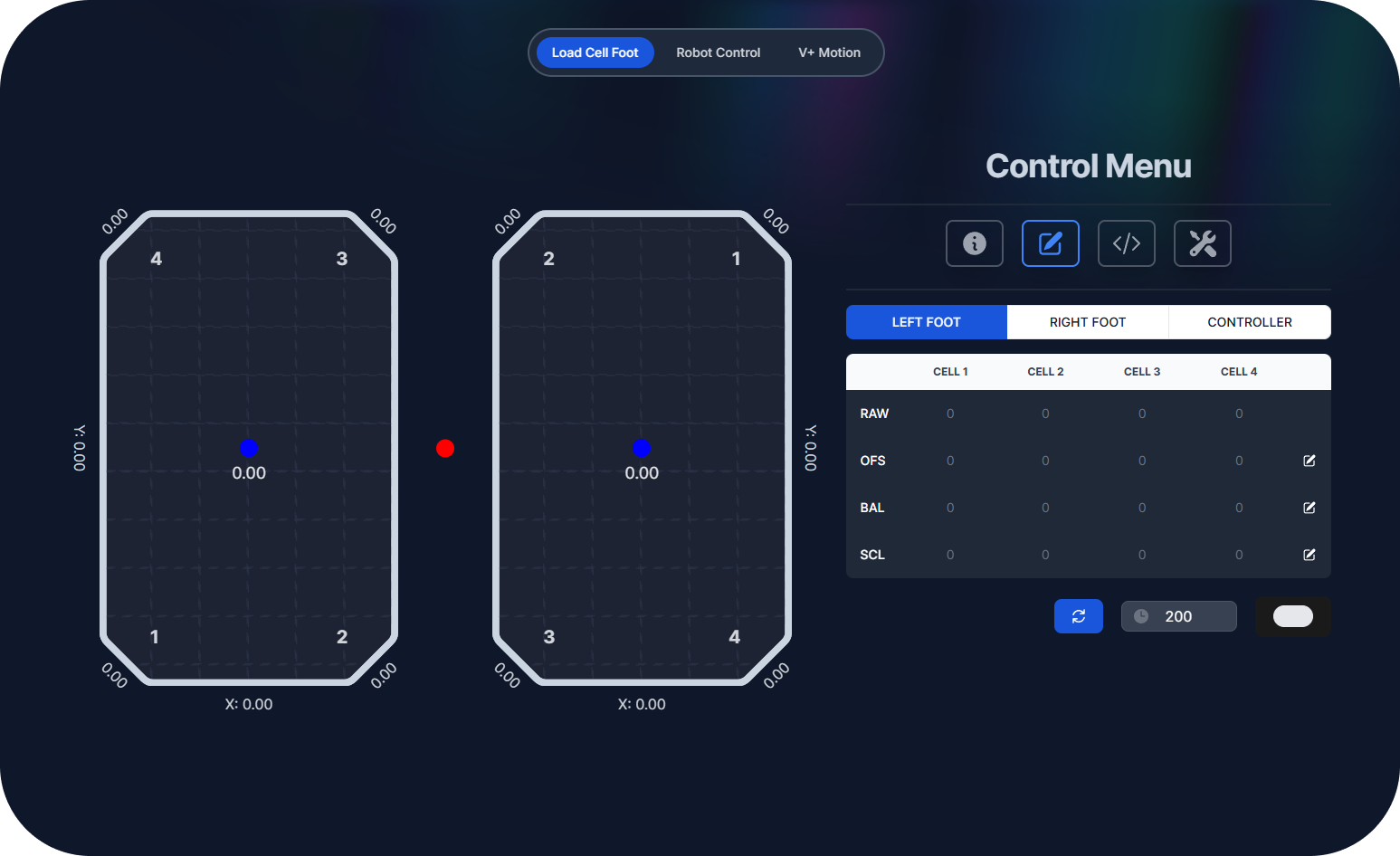}
      \caption{Application Used for Load Cell Configuration}
      \label{fig:Preview_Aplikasi}
    \end{figure}

    \hspace*{1em} Before the load cells can be used, calibration is necessary to measure the load cell coefficients by adjusting parameters such as scaling and offset values. To facilitate configuration, a user-friendly graphical user interface (GUI) application was created, which can be accessed through a browser. This GUI has a control menu for accessing functions and settings and displays the center of pressure to visualize the impact of parameter changes on the system. Figure \ref{fig:Preview_Aplikasi} shows the configuration application interface, including the control menu and the center of pressure display. This application saves configuration data to the EEPROM memory of the microcontroller, so the data remain stored even when the microcontroller is turned off.

\subsection{Center of Pressure Calculation for the Robot}
    \label{subsec:pressurecentercalculation}

    \hspace*{1em} The center of pressure on the robot is calculated by combining data from both robot feet. The main microcontroller communicates with the microcontrollers on the left and right feet to obtain the center of pressure and total pressure data. These data were then used to determine the position of the center of pressure relative to the robot. The center of pressure position on the robot is calculated using Equations \ref{eq:Total_Force_Robot}, \ref{eq:COP_X_Robot}, and \ref{eq:COP_Y_Robot}.
    
    % \begin{figure} [bt] \centering
    %   \includegraphics[width=0.4\textwidth]{gambar/COP_Robot.png}
    %   \caption{Center of Pressure on the Robot (Green), Center of Pressure on the Left Foot (Red), and Center of Pressure on the Right Foot (Blue)}
    %   \label{fig:COP_Robot}
    % \end{figure}

    \begin{align}
        F_{\mathrm{total}} = F_{\mathrm{totalL}} + F_{\mathrm{totalR}}
        \label{eq:Total_Force_Robot} \\
        X_{\mathrm{cop}} = (F_{\mathrm{totalL}} \cdot X_{\mathrm{copL}} + F_{\mathrm{totalR}} \cdot X_{\mathrm{copR}}) / F_{\mathrm{total}}
        \label{eq:COP_X_Robot} \\
        Y_{\mathrm{cop}} = (F_{\mathrm{totalL}} \cdot Y_{\mathrm{copL}} + F_{\mathrm{totalR}} \cdot Y_{\mathrm{copR}})/ F_{\mathrm{total}}
        \label{eq:COP_Y_Robot}  
    \end{align}

The center of pressure data in this study were obtained using a scale. The Y-axis has a scale ranging from -1 to 1. The upper limit was represented by the value 1, whereas the lower limit was represented by the value -1. The X-axis has a scale from -2 to 2, with the right limit represented by the value 2, and the left limit represented by the value -2. For the X-axis, when the robot's weight is supported on one foot, the center of pressure value is positive or negative depending on the position of the supporting foot. When the robot is lifted, the center of pressure value is at (0,0), which is in the middle of the foot.

\subsection{PID Control System Algorithm}
    \label{subsec:algoritmakontrolpid}

    In this study, there are two PID controls: Pitch PID control and Roll PID control. In the Pitch PID control, the input is the position value of the center of pressure on the Y-axis. Meanwhile, in the Roll PID control, the input is the position value of the center of pressure on the X-axis. For the Pitch and Roll PID control setpoints, the values obtained from the previous center of pressure data collection were used. In Equation (\ref{eq:Error_PID}), the error value is obtained from the difference between the current center-of-pressure position and setpoint. Then, in Equation (\ref{eq:Koreksi_PID}), the error value is used to calculate the correction value that will be used to adjust the servo position as compensation.

    \begin{align}
      \mathrm{e} = COP_{\mathrm{error}} = COP_{\mathrm{set}} - COP_{\mathrm{input}}
      \label{eq:Error_PID} \\
      \theta_\mathrm{e} = \mathrm{Kp} \cdot \mathrm{e} + \mathrm{Ki} \cdot \int \mathrm{e} + \mathrm{Kd} \cdot \frac{\mathrm{de}}{\mathrm{dt}}
      \label{eq:Koreksi_PID} 
    \end{align}
    
    The system response stages to the error caused by the difference between the desired and actual center of pressure positions are shown. First, the PID values were set according to the system characteristics, and then the setpoint was set according to the desired center of pressure position. The error was calculated using Equation \ref{eq:Error_PID}. This error value was used to calculate the correction using Equation \ref{eq:Koreksi_PID}. The robot performs movements to maintain balance based on the corrections generated by the PID control, which continues until the robot motion is complete.

    \begin{figure} [bt] \centering
      \includegraphics[width=0.48\textwidth]{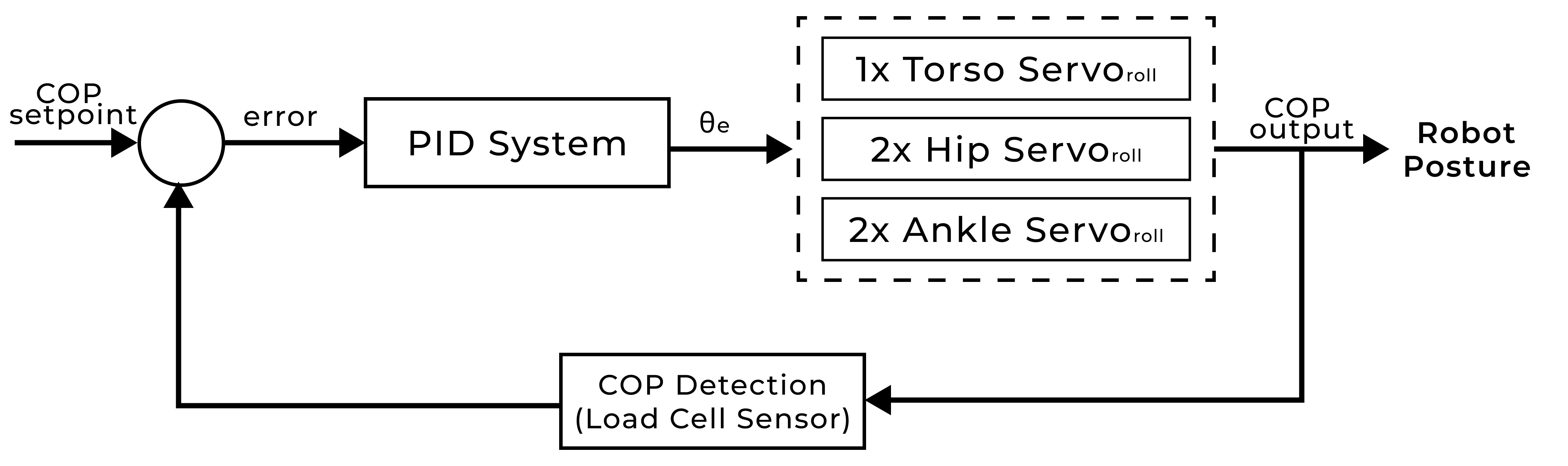}
      \caption{PID Control System Diagram}
      \label{fig:Control_System}
    \end{figure}

    \hspace*{1em} Figure \ref{fig:Control_System} shows the control system diagram consisting of three main blocks: the PID block, the servo block as compensation, and the center of pressure block. The PID System block calculates the correction value using Equation \ref{eq:Koreksi_PID}. The error value was obtained from the difference between $COP_{setpoint}$ and $COP_{input}$. This correction value is used to adjust the servo position in the servo block as a compensation using five servos that control the robot's roll position. The center of pressure block calculates the center of pressure position on the robot's foot and provides this data as an input to the PID control.

\subsection{Servo Settings as Compensation}
    \label{subsec:servosettings}

    \hspace*{1em} In this research, the servos used to maintain balance are the servos that control the robot's roll position. The servos were located on the torso, hip, and ankle. By adjusting several servo angles, the robot can make the necessary adjustments to maintain its balance when moving or standing on uneven surfaces. Five servos were used, consisting of one servo on the torso, two servos on each hip, and two servos on each ankle. The servo settings as compensation can be seen in Figure \ref{fig:Desain_Mekanik} number: 16, 22, 23, 28, 29.

    \begin{align}
        \theta_{\mathrm{torso}} = \theta_{\mathrm{torso}} + (\theta_\mathrm{e}{\mathrm{roll}} \cdot 0.8)
        \label{eq:Koreksi_Torso} \\
        \theta_{\mathrm{hip}} = \theta_{\mathrm{hip}} + (\theta_\mathrm{e}{\mathrm{roll}} \cdot 1)
        \label{eq:Koreksi_Hip} \\
        \theta_{\mathrm{ankle}} = \theta_{\mathrm{ankle}} + (\theta_\mathrm{e}{\mathrm{roll}} \cdot 0.4)
      \label{eq:Koreksi_Ankle}
    \end{align}

In Equation (\ref{eq:Koreksi_Torso}), the correction value is used to adjust the servo position of the torso by a factor of 0.8. In Equation (\ref{eq:Koreksi_Hip}), the correction value is used to adjust the servo position of the hip by a factor of 1. In Equation (\ref{eq:Koreksi_Ankle}), the correction value is used to adjust the servo position of the ankle by a factor of 0.4. The gain coefficients for each joint (0.8 for the torso, 1.0 for the hip, and 0.4 for the ankle) were determined empirically to optimize the balance recovery strategy. A higher gain is assigned to the hip roll joints ($K_{hip}=1.0$) because they have the most significant influence on shifting the robot's Center of Mass (CoM) laterally. Conversely, the ankle gain was kept lower ($K_{ankle}=0.4$) to provide fine postural adjustments without causing foot lift-off, which could destabilize the ground contact.

\section{Result and Discussion}
\label{sec:resultsanddiscussion}

Testing and analysis were performed on the previously designed implementation. The tests included Load Cell Sensor testing on the Robot and the Robot Balance System.

\subsection{Characterization Testing of Each Load Cell}
\label{subsec:results-discussion-characterization}
%\hspace*{1em} 
The load cell sensors were calibrated using five reference masses (50 g, 100 g, 200 g, 500 g, and 1000 g). The smallest weight (50 g) was used to determine the gradient coefficient, whereas the constant was obtained from the tare weight (zero point when the load cell had no load). The resulting data were then used to calculate the error of each load cell by comparing it with the actual mass.

\begin{table}[bt]
\caption{Characterization Results of Load Cell Sensors (Unit: Gram)}
\begin{center}
\begin{tabular}{|c|c|c|c|c|}
\hline
\textbf{Ref. Mass} & \textbf{LC 1} & \textbf{LC 2} & \textbf{LC 3} & \textbf{LC 4} \\
\textbf{(g)} & \textbf{Reading} & \textbf{Reading} & \textbf{Reading} & \textbf{Reading} \\
\hline
50 & 50 & 50 & 50 & 50 \\
100 & 101 & 100 & 103 & 97 \\
200 & 202 & 200 & 203 & 204 \\
500 & 505 & 500 & 494 & 500 \\
1000 & 1004 & 994 & 981 & 1003 \\
\hline
\textbf{Max Error} & \textbf{+5} & \textbf{-6} & \textbf{-19} & \textbf{+4} \\
\hline
\multicolumn{5}{l}{\textit{*LC = Load Cell}}
\end{tabular}
\label{tab:loadcell_combined}
\end{center}
\end{table}
        
The characterization results for each load cell showed measurement errors ranging from 0 to 19 g. These errors were nonlinear, indicating that the measurement errors were not constant across each load cell. Nevertheless, a linear equation can still be used to calculate the actual mass from the load cell readings, although it is not entirely accurate in doing so.

\subsection{Pressure Testing on the Footpads}
    \label{subsec:results-discussion-pressure}

To ensure the accuracy of the Center of Pressure (CoP) calculation defined in Equations (\ref{eq:COP_X_Robot}) and (\ref{eq:COP_Y_Robot}), the input variables, specifically the total force measured on the left foot ($F_{totalL}$) and right foot ($F_{totalR}$), must first be validated. The testing phase aimed to verify the linearity of the load cell readings when subjected to uniform static loads ranging from 0 to 1800 g.

\begin{figure}[bt]
            \centering
            \includegraphics[width=0.48\textwidth]{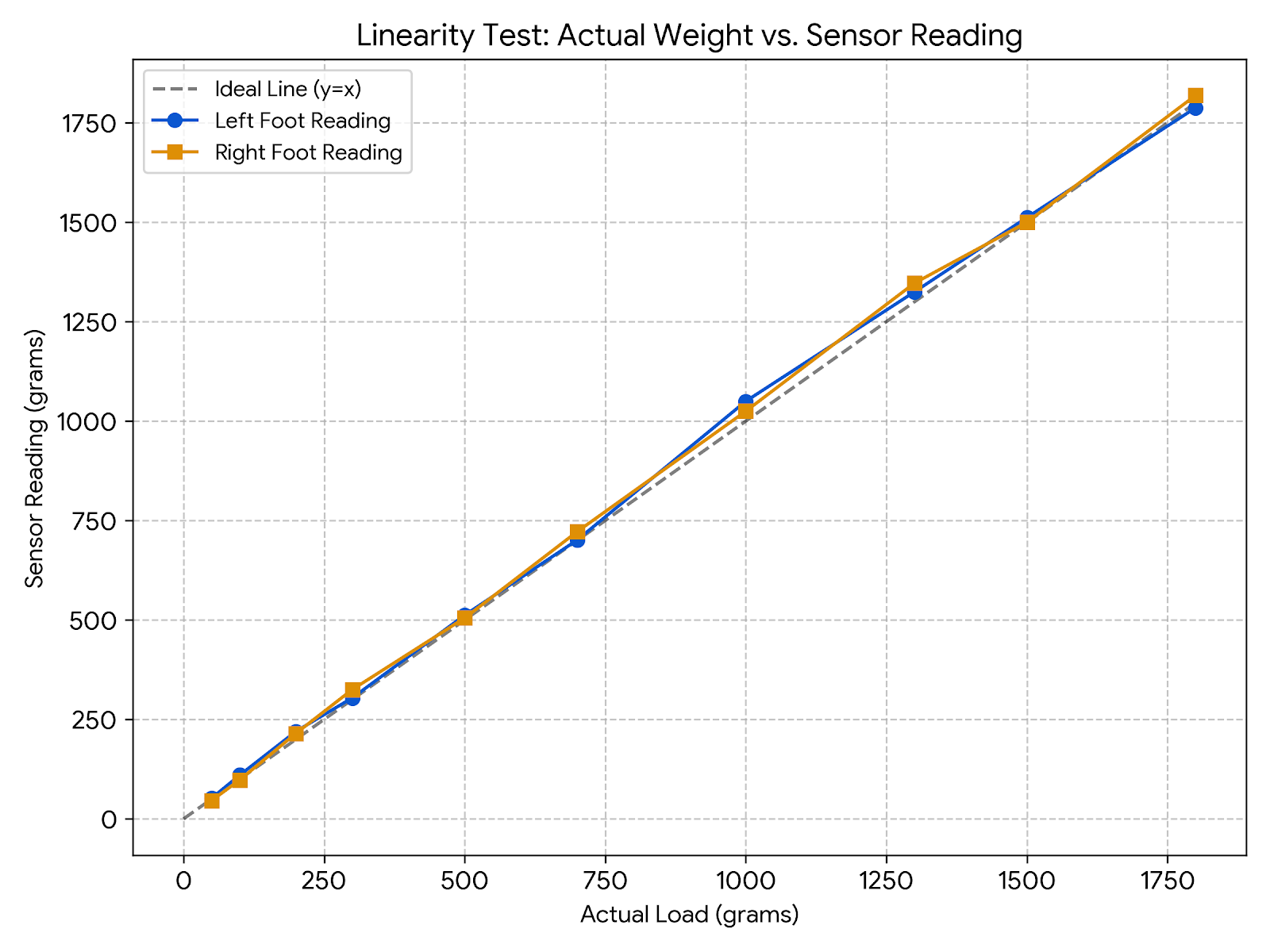}
            \caption{Graph of the Relationship Between Actual Weight and Load Cell Readings for the Left and Right Feet}
            \label{fig:foot_pressure}
        \end{figure}

The measurement results are shown in Fig. \ref{fig:foot_pressure}, demonstrate a linear relationship between the actual mass and the sensor readings for both feet. Although minor deviations were observed, with a maximum error margin of approximately 50 g, the sensor response remained consistent across the loading range. This consistency confirms that the calibrated load cells provide reliable force summation data ($F_{total}$), which serves as the fundamental denominator for the CoP coordinate estimation used in the control system.

\subsection{Center of Pressure Testing on the Robot}
    \label{subsec:hasil-pembahasan-pusat-tekanan}
        
        The test was performed while the robot moved and data were collected from the sensor every 50 ms. The goal was to obtain the center of pressure data during the robot's movements. The test involved collecting three data points as the robot walked in place: when it was in the double-support position, when it lifted the right foot, and when it lifted the left foot.
        
        A picture of the robot foot was used to show the center of pressure data from the robot. The center of pressure on the X and Y axes is represented by points with a 2:1 scale. This image helps us to determine the center of pressure when the robot walks in place. The data were obtained from three center of pressure points collected earlier.
        
\begin{figure}[t] 
    \begin{subfigure}[b]{0.5\textwidth}
        \centering
        \includegraphics[width=0.8\textwidth]{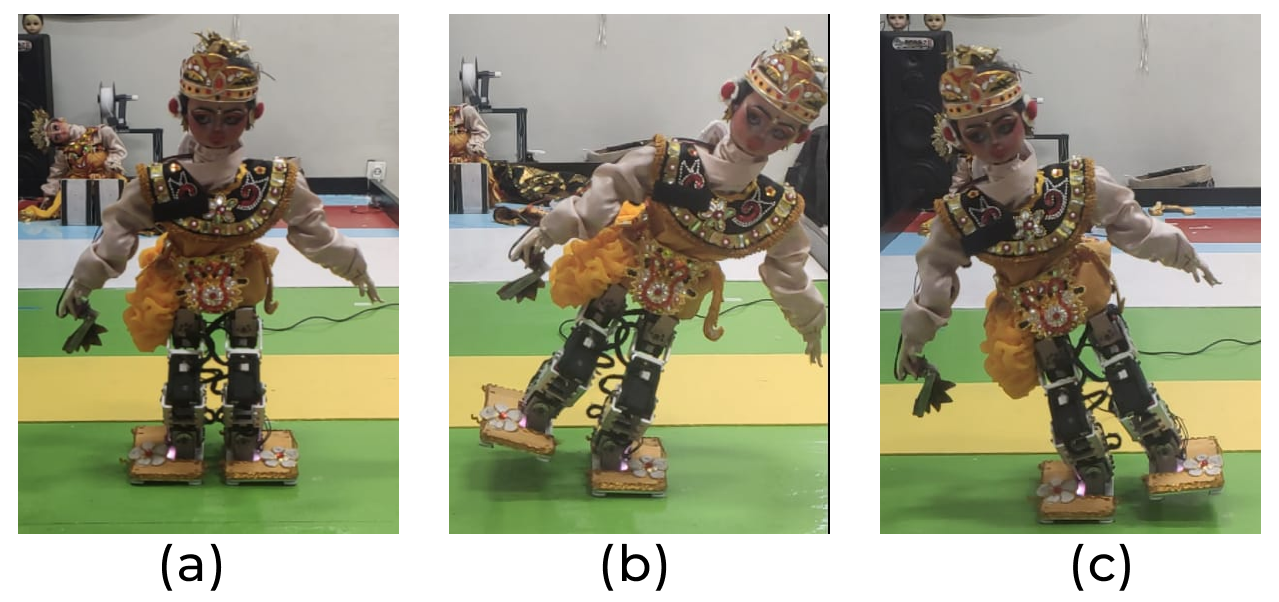}
        \label{fig:Ng1} 
    \end{subfigure}

\medskip % insert a bit of vertical whitespace
    \begin{subfigure}[b]{0.48\textwidth}
        \centering
        \includegraphics[width=\linewidth]{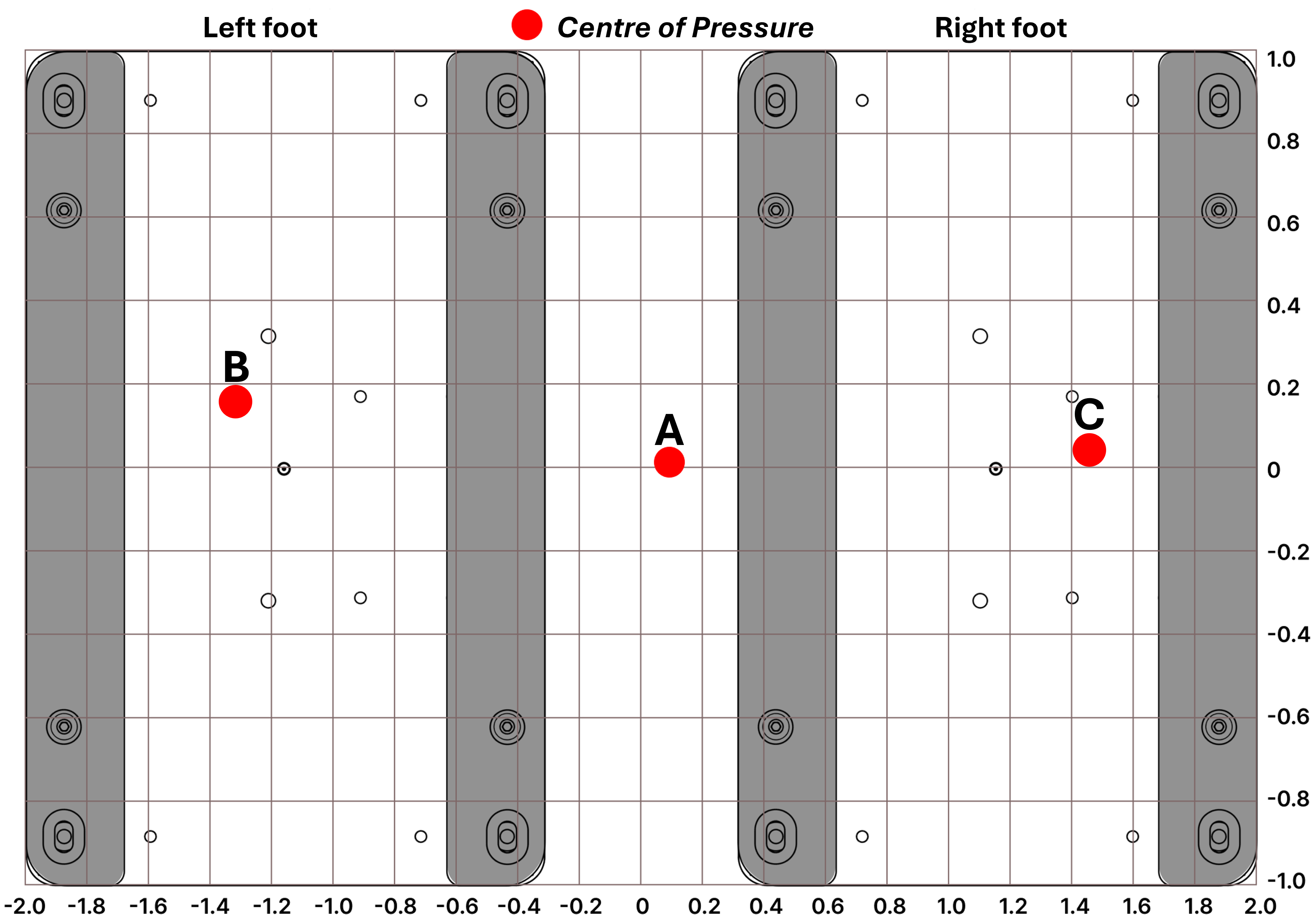}
    \label{fig:Ng2}
    \end{subfigure}

    \caption{Visualization of the Center of Pressure (CoP) in relation to the robot's feet during walking in place. (a) CoP is in the middle when both feet are on the ground. (b) CoP moves to the left foot when the right foot is lifted. (c) CoP moves to the right foot when the left foot is lifted.}
    \label{fig:cop_combined_analysis}
\end{figure}

        \hspace*{1em} In Figure \ref{fig:cop_combined_analysis} (a), the center of pressure on the X-axis is 0.09 and on the Y-axis is 0.01. In Figure \ref{fig:cop_combined_analysis} (b), the center of pressure on the X-axis is -1.30, and on the Y-axis is 0.16. Meanwhile, in Figure \ref{fig:cop_combined_analysis} (c), the center of pressure on the X-axis is 1.44, and on the Y-axis is 0.03. From this visualization, the center of pressure on the X-axis changes significantly when the robot lifts the left and right feet, whereas on the Y-axis, the center of pressure coordinates remain in the middle of the foot.

\subsection{PID Control System Testing}
    \label{subsec:hasil-pembahasan-pid}
        
\begin{figure}[bt]
            \centering
            \includegraphics[width=0.4\textwidth]{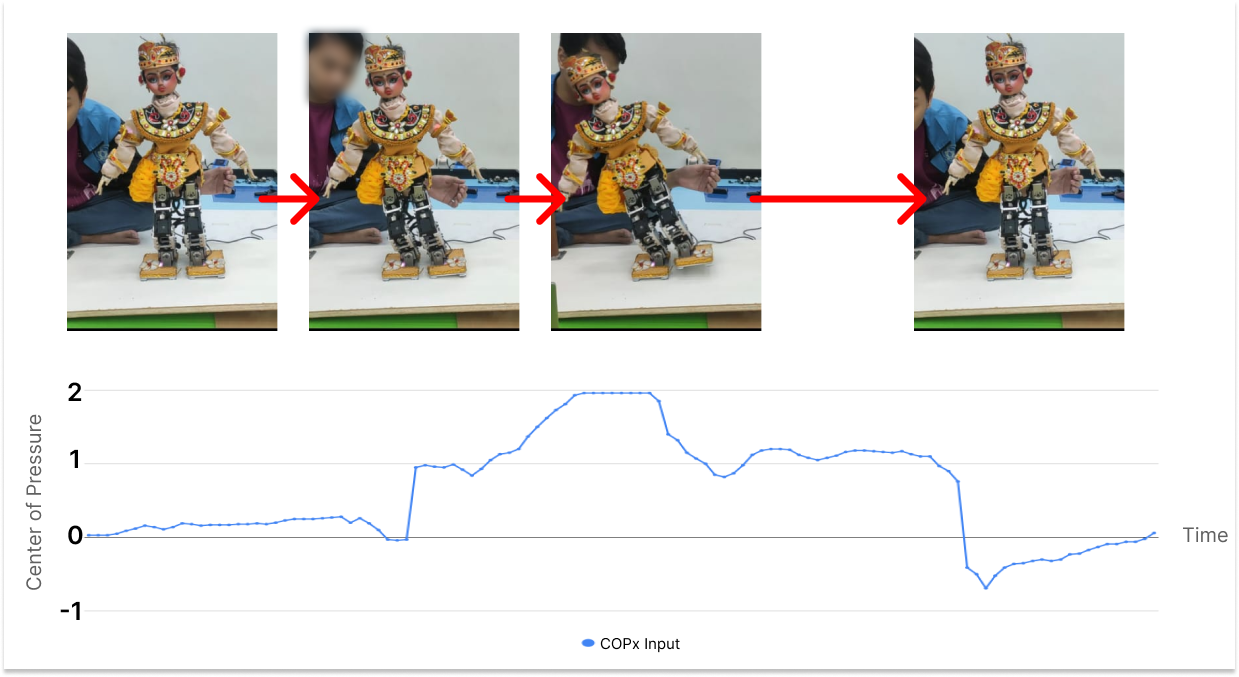}
            \caption{System Response Graph With PID Control (Not Fall)}
            \label{fig:robot_not_fall}
        \end{figure}

\begin{figure}[bt]
            \centering
            \includegraphics[width=0.4\textwidth]{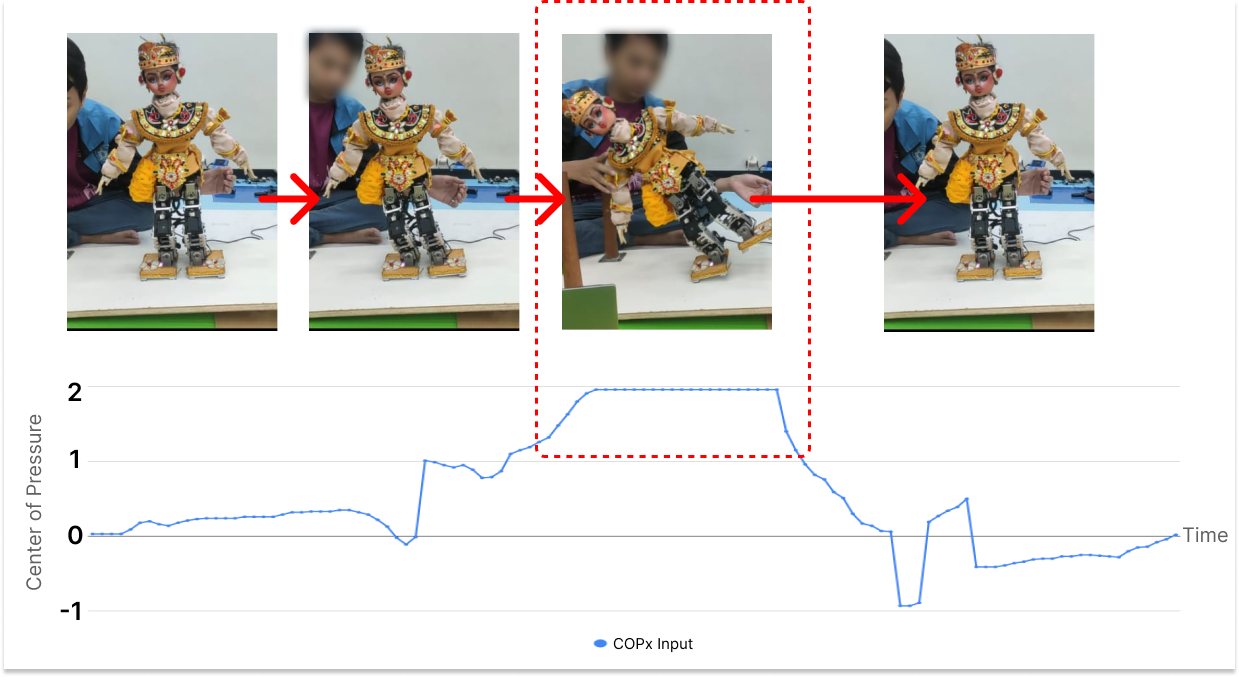}
            \caption{System Response Graph Without PID Control (Fall)}
            \label{fig:robot_fall}
        \end{figure}

        \hspace*{1em} In this test, the robot's balance system was tested using PID control. The results to be analyzed were the system response when using PID control and without PID control. The test was conducted by lifting the right or left foot at a 3-degree tilt. Figure \ref{fig:robot_not_fall} shows the test results obtained using the PID controller. Figure \ref{fig:robot_fall} shows the results without the PID controller. As shown in Figure \ref{fig:robot_not_fall}, if the center of pressure value on the X-axis exceeds the maximum limit within a certain time, the robot will fall. This indicates that the error value produced by the PID controller cannot be calculated accurately because the input value received by the PID controller does not match the expected value.

\subsection{Influence of PID Parameters Testing}
    \label{subsec:hasil-pembahasan-parameter-pid}
        
        In this test, we examined the effect of each PID parameter on the system. We checked how these parameters changed the system's response and the Root Mean Square (RMS) error values. The test was performed as before, by tilting the right or left foot by 3 °.
        
        \begin{table}[bt]
            \centering
            \caption{Effect of Parameter $K_p$ on Right Foot Lifting with 3-Degree Tilt}
            \begin{tabular}{|c|c|c|c|c|}
                \hline
                \textbf{PID} & \textbf{Fall} & \textbf{Not Fall} & \textbf{Success} & RMS Error \\
                \hline
                $K_p = 0.00$ & 6 & 0 & 0   \% & 0.7598 \\
                $K_p = 0.05$ & 6 & 0 & 0   \% & 0.7690 \\
                $K_p = 0.10$ & 0 & 6 & 100 \% & 0.7779 \\
                $K_p = 0.15$ & 1 & 5 & 83  \% & 0.8145 \\
                $K_p = 0.20$ & 1 & 5 & 83  \% & 0.8870 \\
                $K_p = 0.25$ & 0 & 6 & 100 \% & 0.8801 \\
                \hline
            \end{tabular}
            \label{tab:testing_p}
        \end{table}

The results in Table \ref{tab:testing_p} show that the best $K_p$ value for maintaining the balance of the robot is between 0.10 and 0.20. When $K_p$ was 0.00 or 0.05, the robot always fell over. The $K_p = 0.10$ value worked best, with the robot remaining balanced 100\% of the time when lifting its right or left foot at a 3-degree tilt. Higher $K_p$ values, such as $K_p = 0.25$, worsened the performance. The Root Mean Square (RMS) results show the smallest error at $K_p = 0.10$, with a value of 0.7779. This shows the importance of choosing the appropriate $K_p$ value for robot stability.

    \begin{table}[bt]
        \centering
        \caption{Effect of Parameter $K_i$ on Right Foot Lifting with 3-Degree Tilt}
        \begin{tabular}{|c|c|c|c|c|}
            \hline
            \textbf{PID} & \textbf{Fall} & \textbf{Not Fall} & \textbf{Success} & RMS Error \\
            \hline
            $K_p = 0.1, K_i = 0.01$ & 2 & 4 & 66 \%  & 0.9701\\
            $K_p = 0.1, K_i = 0.02$ & 2 & 4 & 66 \%  & 0.8950\\
            $K_p = 0.1, K_i = 0.04$ & 1 & 5 & 83 \%  & 0.9345\\
            $K_p = 0.1, K_i = 0.10$ & 0 & 6 & 100 \% & 0.8471\\
            $K_p = 0.1, K_i = 0.20$ & 0 & 6 & 100 \% & 0.8980\\           
            \hline
        \end{tabular}
        \label{tab:testing_pi}
    \end{table}

The results in Table \ref{tab:testing_pi} show that the optimal $K_i$ value for maintaining the balance of the robot is between 0.10 and 0.20. When $K_i$ was 0.01 or 0.02, the robot did not balance well, with a success rate of 66\% to 83\%. However, at $K_i = 0.10$ and $K_i = 0.20$, the robot was balanced perfectly, with a 100\% success rate when lifting the right or left foot at a 3-degree tilt. The Root Mean Square (RMS) results were lowest at $K_i = 0.10$ with 0.8471, and highest at $K_i = 0.01$ with 0.9701. The $K_i$ value does not significantly affect the performance, and sometimes higher $K_i$ values worsen the performance. This implies that the $K_i$ parameter in the PID control is not very important for this system. Although increasing $K_p$ to 0.25 maintained a 100\% success rate, the RMS error was higher than that with the best $K_p=0.10$. This implies that higher proportional gains cause overshoot and low-frequency oscillations around the setpoint. The robot reacted too strongly to small CoP changes, leading to a less stable posture, as indicated by higher RMS error values.

    \begin{table}[tb]
        \centering
        \caption{Effect of Parameter $K_d$ on Right Foot Lifting with 3-Degree Tilt}
        \begin{tabular}{|c|c|c|c|c|}
            \hline
            \textbf{PID} & \textbf{Fall} & \textbf{Not Fall} & \textbf{Success} & RMS Error \\
            \hline
            $K_p = 0.1, K_d = 0.005$ & 0 & 6 & 100 \% & 0.7143 \\
            $K_p = 0.1, K_d = 0.010$ & 0 & 6 & 100 \% & 0.7077 \\
            $K_p = 0.1, K_d = 0.020$ & 2 & 4 & 66  \% & 0.7262 \\
            $K_p = 0.1, K_d = 0.050$ & 5 & 1 & 16  \% & 0.7344 \\
            $K_p = 0.1, K_d = 0.100$ & 6 & 0 & 0   \% & 0.9231 \\          
            \hline
        \end{tabular}
        \label{tab:testing_pd}
    \end{table}

The results shown in Table \ref{tab:testing_pd} indicate that the optimal $K_d$ parameter value for maintaining the robot balance ranged from 0.005 to 0.020. At $K_d = 0.005$ and $K_d = 0.010$, the robot successfully maintained its balance with a 100\% success rate. However, at $K_d = 0.020$, the success rate decreases to 66\%, and at $K_d = 0.050$, the success rate is only 16\%. At $K_d = 0.100$, the robot consistently fell over. The Root Mean Square (RMS) results show the lowest value at $K_d = 0.010$ with 0.7077, while the highest value at $K_d = 0.100$ is 0.9231. While the derivative term ($K_d$) typically dampens oscillations, the results in Table \ref{tab:testing_pd} show that excessive $K_d$ values (e.g., $K_d=0.100$) cause total system failure (0\% success rate). This phenomenon can be attributed to the noise amplification. Because load cell readings contain inherent high-frequency measurement noise (as characterized in Section IV-A), a high derivative gain amplifies these fluctuations, causing the servos to jitter violently. This mechanical instability prevents the robot from maintaining the Center of Pressure (CoP) within the support polygon, leading to falls.

\section{Conclusion}
In this study, a balance system based on load cells was developed for a humanoid robot with 29 degrees of freedom, specifically for the VI-ROSE ITS humanoid dance robot. The system, embedded in the soles of the feet and using the ESP32-C3 microcontroller to read the load cell sensors, was successfully created with measurement errors ranging from 0-19 grams per load cell and 0-50 grams for the total foot. The average errors recorded were 14.8 g for the right foot and 14.6 g for the left foot. The robot successfully balanced itself while lifting either the right or left foot by applying PID control to five servos that adjusted the robot's roll position, including one servo in the torso, two servos in the hip, and two servos in the ankle. Manual PID tuning results showed that $K_p = 0.1$ and $K_d = 0.005$ provided the best performance, achieving a 100\% success rate in maintaining the robot balance. Testing was performed using foot-lifting movements at a 3-degree tilt.

For the future development of this system, it is recommended to replace the servos with more robust models, such as MX-64T, for the motors that control the robot's roll position. Additionally, using the Center of Pressure (CoP) data to determine the zero-moment point (ZMP) could help improve the robot's balance accuracy.

%\newpage

\section*{Acknowledgment}
This research is funded by the Indonesian Endowment Fund for Education (LPDP) on behalf of the Indonesian Ministry of Higher Education, Science and Technology and managed under the EQUITY Program (Contract No 4299/B3/DT.03.08/2025 \& No 3029/PKS/ITS/2025)

\end{document}